# Nordic Vehicle Dataset (NVD): Performance of vehicle detectors using newly captured NVD from UAV in different snowy weather conditions.

Hamam Mokayed
LTU
Luleå, Sweden
Hamam.mokayed@ltu.se

Amirhossein Nayebiastaneh
LTU
Luleå, Sweden
aminay-2@student.ltu.se

Kanjar De
LTU
Luleå, Sweden
kanjar.de@associated.ltu.se

Stergios Sozos
LTU
Luleå, Sweden
stesoz-2@student.ltu.se

Olle Hagner
Smartplanes
Jävre, Sweden
olle.hagner@smartplanes.se

Björn Backe
LTU
Luleå, Sweden
bjorn.backe@ltu.se

## Abstract

*Vehicle detection and recognition in drone images is a complex problem that has been used for different safety purposes. The main challenge of these images is captured at oblique angles and poses several challenges like non-uniform illumination effect, degradations, blur, occlusion, loss of visibility, etc. Additionally, weather conditions play a crucial role in causing safety concerns and add another high level of challenge to the collected data. Over the past few decades, various techniques have been employed to detect and track vehicles in different weather conditions. However, detecting vehicles in heavy snow is still in the early stages because of a lack of available data. Furthermore, there has been no research on detecting vehicles in snowy weather using real images captured by unmanned aerial vehicles (UAVs). This study aims to address this gap by providing the scientific community with data on vehicles captured by UAVs in different settings and under various snow cover conditions in the Nordic region. The data covers different adverse weather conditions like overcast with snowfall, low light and low contrast conditions with patchy snow cover, high brightness, sunlight, fresh snow, and the temperature reaching far below – 0 degrees Celsius. The study also evaluates the performance of commonly used object detection methods such as YOLOv8s, YOLOv5s, and Faster RCNN. Additionally, data augmentation techniques are explored, and those that enhance the detectors' performance in such scenarios are proposed. The code and the dataset will be available at https://nvd.ltu-ai.dev*

## 1. Introduction

In the Arctic region of Scandinavia, drones are used for monitoring purposes in search and rescue missions. Drones can be first on-site when accidents occur like car accidents or traffic congestion to provide an overview of the event scene which can be lifesaving. In rural areas, long distances between cities and villages and harsh weather conditions such as snow, snow fog, snowstorms, and temperatures reaching far below – 0 degrees Celsius make search and rescue missions by drones difficult. During wintertime the light conditions in the northern hemisphere are low, and for upper northern Scandinavia during the occurrence of Polar night, the sun never rises above the horizon. During wintertime traffic monitoring for road maintenance purposes with drones is a timesaving and more environmentally friendly option than using cars or trucks to inspect the roads. Monitoring bottlenecks in traffic in more urban areas is also of interest to early drivers commuting to work. In a snowy landscape with snowy cars and low light conditions, it is difficult to detect cars from the air, even by the human eye. Detecting objects concealed by snow presents unique challenges compared to other scenarios, primarily because most existing detectors are trained on datasets that either contain images captured under normal weather conditions [1-2] or on artificially generated snow images [3-5]. However, these models are not effective in detecting objects in snowy conditions since snow hides many of the visual features highly crucial and required for object detection.

This paper assesses how well object detectors perform using a dataset captured by unmanned aerial vehicles (UAVs) in various winter weather conditions, ranging from light to complete snow cover. The goal is to investigate whether detectors perform poorly in such conditions and to highlight the importance of using adequate training datasets when developing detectors. The primary focus of this study is using UAV images to detect vehicles captured in a wide range of winter weather conditions with various degrees of snow cover and not limited to roads. To understand the novelty and uniqueness of our approach, we conducted an extensive search for research papers or projects with a scope like ours. We found datasets that use UAV images for vehicle detection but have significant differences from our collected dataset, leading to different scopes and challenges. In the following section, we will attempt to provide a technical summary of other datasets captured by drones that have been used in other research.

## 2. UAVs dataset

In this section, we will analyze each of the research papers and projects that use images captured by UAV and compare their datasets to ours with the aim of highlighting the key differences that make our research stand out. We aim to demonstrate the novelty and contribution of our research in the field of vehicle detection using UAV images in different weather conditions such as heavy snow.

**VisDrone dataset** [6-8]: The need for computer vision in analyzing visual data collected from drones has led to the creation of a comprehensive benchmark dataset called VisDrone. Developed in China, this dataset was intended to facilitate various computer vision tasks related to drone imagery. The VisDrone2019 dataset represents an effort to merge the fields of computer vision and drone technology, however, emphasis is given to object detection regardless of the weather conditions and is not limited to vehicle detection. It does contain cars, but it also includes other kinds of objects such as pedestrians, bicycles, etc. Thus, it is expected that the models built on top of this dataset are not specialized in vehicle detection under extreme weather conditions. It is also based on a different continent, which can have a very different view from a drone compared to a European city. The benchmark dataset contains 288 video clips and 10,209 static images captured by drone-mounted cameras in various locations, environments, objects, and densities in China. The dataset was collected using different drone models, scenarios, and weather and lighting conditions. The frames are manually annotated with more than 2.6 million bounding boxes of objects of interest.

**UAV project** [9-11]: This dataset was slightly more like our perspective than the others. It was designed to be a challenging dataset for existing object detection solutions trained on limited datasets. While it was intended to include various weather conditions, its distribution of weather conditions suggests that it only includes fog as an adverse weather condition. The vehicles annotated in this dataset are also present only on the road, which is again a main difference from our dataset. It is also mentioned that in some places, the vehicles were too small to classify them or assess their motion, which is a key difference from our dataset, which aims to identify all vehicles, regardless of their size, if it is identifiable by us to annotate. The dataset consists of 10 hours of raw video that make up the proposed UAVDT benchmark and was cut down to 100 sequences with roughly 80,000 representative frames each. The sequences range in frame count from 83 to 2970. A UAV platform was used to make films in a variety of metropolitan settings, including squares, highways, crossings, toll booths, arterial routes, and T-junctions. The video sequences are captured at 30 frames per second (fps) and in a 1080 x 540-pixel resolution. In the dataset, which included 2,700 automobiles, around 80,000 frames from the 10 hours of raw footage were annotated with 0,84 million bounding boxes.

**UAV-Vehicle-Detection-Dataset** [12-13]: This dataset was created to address the orientation and scale-invariant problem, with a focus on detecting and re-identifying vehicles. However, it differs from our research in that it is primarily concerned with identifying vehicles on roads, while our dataset and research aim to identify vehicles in any location. Additionally, the dataset only includes images captured under normal weather conditions without any adverse weather conditions such as rain or snow. There is a similarity between this dataset and ours in terms of capturing vehicles from various angles, resulting in significant perspective distortions, but this is common in most UAV datasets. The training dataset for the vehicle detector is generated from 3 different sources. It consists of 154 images from the aerial-cars-dataset in GitHub, which comes from a video with no extreme weather conditions, 1374 images from the UAV-benchmark-M, and a dataset of 157 custom labeled images. The proposed solution for live tracking of vehicles by detection approach is using 11 frames per second on color videos of 2720p resolution to perform in an efficient way.

**Mimos drone dataset** [14-15]: This paper starts with the same motivation of our paper, which is protecting and securing specific areas by text detection. However, it follows a different path, by aiming at identifying the text on the plates of the cars, or any other text on the car. Thus, although the initial motivation is similar, the dataset used and the research itself is completely different from our scope. Added to that, weather conditions are not considered at all during this research, which plays a key role in our research. The dataset consists of 1142 images and most of the photos contain parking signs or traffic signals. This work focuses on low altitude captured images, the ranges are from 1-3, 3-5, and more than 7 meters at different angles. As a result, the dataset contains photos with tiny text and license plate numbers that are in low resolution.

**Data synthesizing** [16]: This paper had been a great addition to our research, as it is highlighting the lack of datasets containing adverse weather conditions. Its aim is to build a model that generates rendered weather conditions on images which is different from our scope, but the approach that is followed has some interesting key points for us. Two datasets are used for training the model, the Flickr Weather Image Dataset, and the CARISSMA Weather Image Dataset. Both contain images that are not recorded from UAV, but they both contain different weather conditions such as fog, rain, or snow, and they also both contain car objects. The datasets are only focusing on street videos, and since they are not recorded from UAV the angle is completely different. Thus, even though this research tends to have some similarities to our challenges, it is still completely different.

**UAV videos for traffic** [17]: This research's technical implementation is closer to our general goal, which is detecting vehicles through UAV data. However, the scope of this research is to build a model to get traffic information, which means that vehicle detection is restricted to cars on the streets. Also, no context is given about the weather conditions, which is the challenge we aim to explore and solve. The data was captured by UAV on the main city roads of Chongqing, at a height of 200-250 meters above the ground. The video has a high resolution of 3840 pixels by 2160 pixels. The data was created using the VOC2007 standard.

**Other available UAV Dataset**: As there are many other datasets used for vehicle detection, we will try to list some of them in the following table and crossmatch them with ours based on the following criteria as explained in both table 1 and figure 1.

- Criteria 1 (C1): Data captured from UAV.
- Criteria 2 (C2): Data has vehicles.
- Criteria 3 (C3): Location is generic.
- Criteria 4 (C4): Varying weather conditions.

Table 1. Applied search criteria over available vehicle dataset.

| Dataset | C1 | C2 | C3 | C4 | Additional info |
|---|---|---|---|---|---|
| DLR 3K [17] | ✗ | | | | |
| VEDAI-512 [18] | ✗ | | | | |
| VEDAI-1024 [19] | ✗ | | | | |
| DOTA [20] | ✓ | ✓ | ✗ | ✗ | Aerial Images from different platforms, not specific from UAV, contains vehicles, but no information about places of images (exclusive on streets, or not) or weather conditions |
| Stanford Drone [21] | ✓ | ✓ | ✓ | ✗ | Focused on campus area, which is a special case, with very specific type of vehicles, and very constrained space. |
| CARPK [22] | ✓ | ✓ | ✗ | ✗ | Only restricted to parking lots |
| PUCPR+ [22] | ✗ | | | | |
| CyCAR [23] | ✓ | ✓ | ✗ | ✗ | |
| UA123 [24] | ✓ | ✓ | ✗ | ✗ | |
| UAVDT [25] | ✓ | ✓ | ✗ | ✓ | No extreme weather conditions included |
| MOR-UAV [26] | ✓ | ✓ | ✗ | ✓ | It contains different scenarios such as nighttime, occlusion and camera motion |
| BIRDSAI [27] | ✓ | ✗ | | | |
| MOHR [28] | ✓ | ✓ | ✗ | ✗ | |
| NVD | ✓ | ✓ | ✓ | ✓ | Different weather conditions with different snow levels |

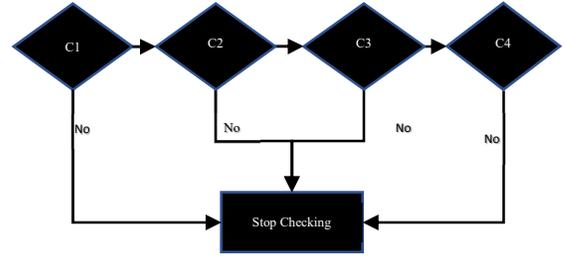

Figure 1. sequential checking search criteria

## 3. Nordic vehicle dataset (NVD)

### 3.1. Data Capturing

The video datasets were acquired using a Freya unmanned aircraft system (figure 2). The flights were conducted autonomously according to preprogrammed flight plans at altitudes varying from 120 m up to 250 m above ground level. The Freya unmanned aircraft specifications are explained in table2.

Table 2. Specification of Freya unmanned aircraft.

| Airframe type | Flying wing |
|---|---|
| Propulsion | Electric, pusher propeller |
| Wing span | 120 cm |
| Take-off weight | 1.2 kg |
| Cruise speed | 13 m/s (47 km/hr) |

While the specifications of the camera used to capture the image are shown in table 3.

Table 3. Specification of Freya unmanned aircraft.

| Sensor | 1.0-type (13.2 x 8.8 mm) Eximor RS CMOS |
|---|---|
| Lens | f=7.9 mm (35 mm format equivalent 24mm), F 4.0 |
| Video recording | 1080 p or 4K at 25 frames per second |
| Still image | 16 Mpix |
| Sensor | 1.0-type (13.2 x 8.8 mm) Eximor RS CMOS |

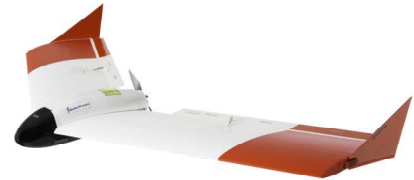

Figure 2. Freya unmanned aircraft.

### 3.2. Data preparation

**Data annotation**: The CVAT tool [29] was used in both online and self-hosted server setups to annotate the captured images and videos. CVAT provides rectangular bounding boxes for each object in a format that can be used by different detectors.

**Data augmentation**: Data augmentation is a crucial technique for improving the performance of object detection models. By increasing the effective size of the dataset, data augmentation helps prevent overfitting and improve generalization.

At the same time, it helps to create a more diverse training dataset, exposing the model to a wider range of examples. Here lies the reason, on why we used albumentations. In our case, we aim on a wide range of weather conditions. We employed the technique of albumentation for weather simulation as some of our data that had normal weather conditions. The impact of using albumentation library over the NVD is evaluated in Sec 4. we initially used the albumentations library to augment our data for training and testing. We applied pixel-level transformations, such as simulating weather conditions like snow, rain, and fog. To keep track of the augmented frames, we saved each modified image with a unique identifier. We also made sure that any bounding boxes we had in the original images remained accurate in the augmented image. To do this, we created new annotations that replicated the original bounding boxes and added the unique identifier to the filename. This method was offline which required enormous disk space and processing time. This restriction led us to using the YOLO built-in augmentation which is implemented online. YOLO needs hyperparameters to define different configurations that impact the model training process. Therefore, we assigned values to the hyperparameters that influence data augmentation, which helps to improve our dataset during training. Some hyperparameters that we have set, which affect data augmentation, are listed below, but the entire set can be accessed through the code available on Github.

- fl_gamma: 0.0  # focal loss gamma.
- hsv_h: 0.015   # image HSV-Hue augmentation (fraction)
- hsv_s: 0.7 # image HSV-Saturation augmentation (fraction)
- degrees: 45.0  # image rotation (+/- deg)

**Flight height estimation**: As part of the data classification process, we used flight height as one of the factors. To estimate the flight height, we developed a method that utilizes the diagonal length of the bounding box in a frame, which we obtained from annotation data, perspective geometry, and maximum flight height information from UAV. The accuracy of the applied method to estimate the height of the UAV is explained in figure 3.

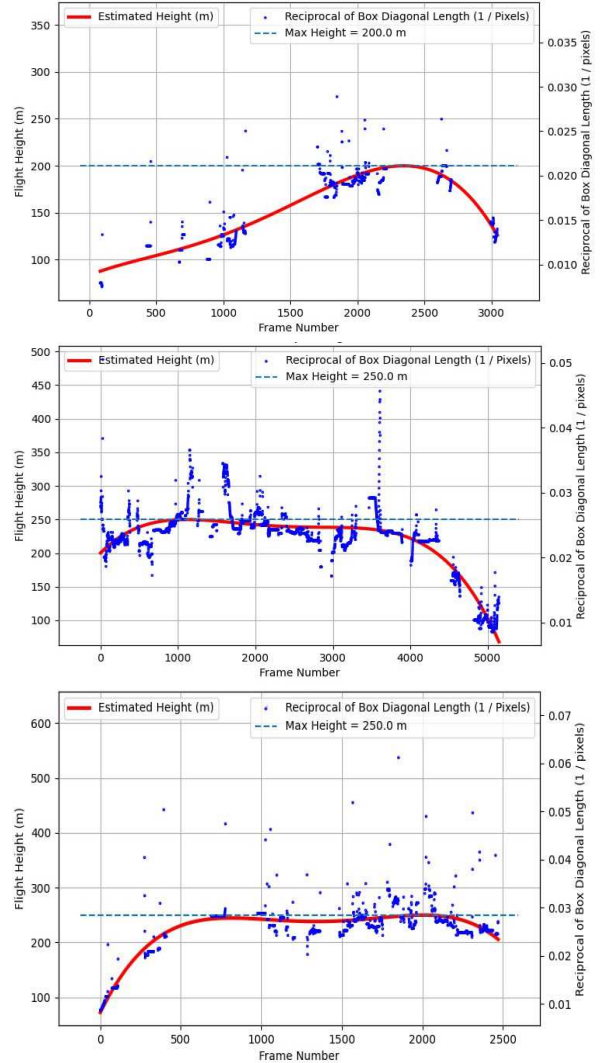

Figure 3. Estimation of flight altitude for different videos.

The size of the bounding box in an aerial image can be used as an indicator of the flight height. In this work, we use the diagonal length of the bounding box denoted by $l$, as the vehicle's size in the image plane in pixels. We assume that the diagonal length of the vehicle is denoted by $L$ (in meters) in the real world.

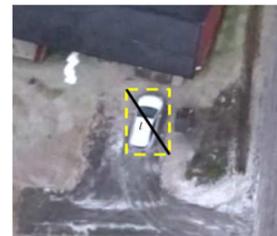

Figure 4. Bounding box's size defined by its diagonal length $l$.

Figure 4 illustrates the diagonal length of the bounding box ($l$) in an aerial image. $l$ is calculated as the Euclidean distance between the top-left and bottom-right corners of the bounding box. To estimate the flight height, we need to establish a relationship between $l$ and the flight height $H$ (in meters). Figure 5 depicts the perspective geometry of a camera placed at height of $H$, observing an object with a real diagonal length of $L$ meters. In the figure the focal length of the camera is denoted by $f$.

$$H = fL\lambda$$
$$\lambda = \frac{1}{l}$$

This relationship will be utilized to estimate the flight height in each frame of the video. We assume that f is constant, and L is the same for each vehicle. In each frame, we set l as the mean bounding box size among all vehicles, and H will be proportional to $\lambda$. To obtain a polynomial fit for each frame, we fit a fourth-degree polynomial to the values of $\lambda$. However, the above simplifications result in numerous outliers, to handle them. After fitting the polynomial, we determine its maximum value $\lambda_{max}$. Since we know the maximum flight height $H_{max}$ for each video, we can calculate the flight height for each frame of the video using the formula:

$$H = \frac{H_{max}}{\lambda_{max}}\lambda$$

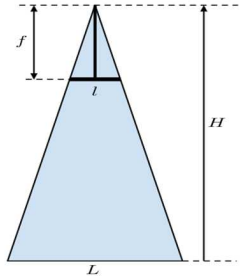

Figure 5. Perspective geometry of a camera placed at height H observing an object with real diagonal length of L.

### 3.3. Data Description

Nordic vehicle dataset (NVD) comprises 22 videos of aerial footage captured in the north of Sweden, with mostly snowy weather conditions. The flight altitudes range from 120 to 250 meters, with varying snow cover and cloud cover. The annotated videos have a total of 8450 annotated frames, containing 26313 annotated cars. The resolution of the videos varies from 1920 x 1080 to 3840 x 2160, with a frame rate of 5 or 25 frames per second. The GSD (Ground Sample Distance) or pixel size ranges from 11.1 cm to 22.2 cm, with some videos being stabilized to ensure smoother footage. Overall, our dataset provides a diverse collection of aerial images of cars in snowy conditions in northern Sweden, with annotated data that can be utilized for various applications concerning safety in the region. The following image samples were taken from various videos under different conditions, and the vehicles have been enlarged for better illustration.

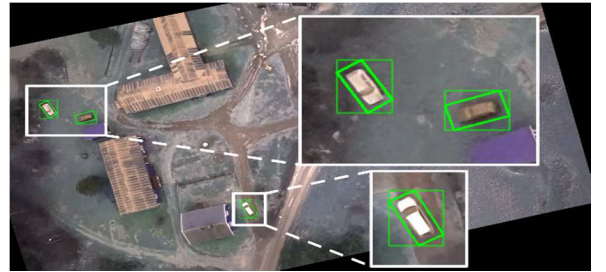

Sample1: altitude (150), Snow cover (0-1cm), cloud cover (overcast).

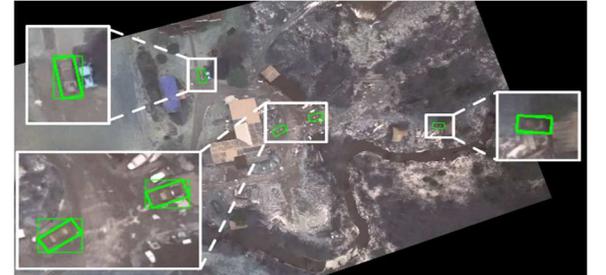

Sample2: altitude (150), Snow cover (0-1cm), cloud cover (overcast).

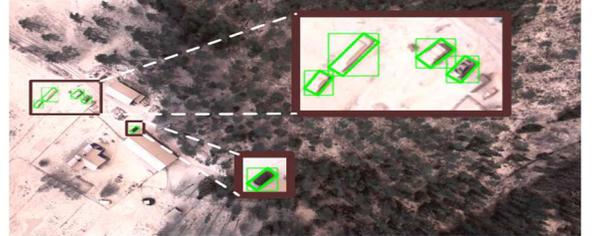

Sample3: altitude (250), Snow cover (1-2cm), cloud cover (light).

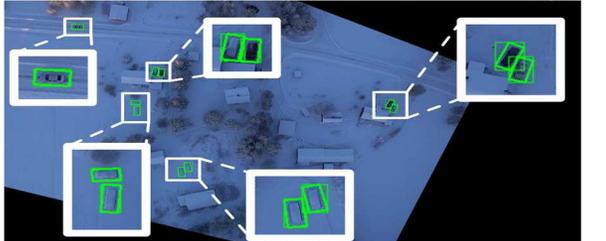

Sample4: altitude (250), Snow cover (5-10), cloud cover (clear).

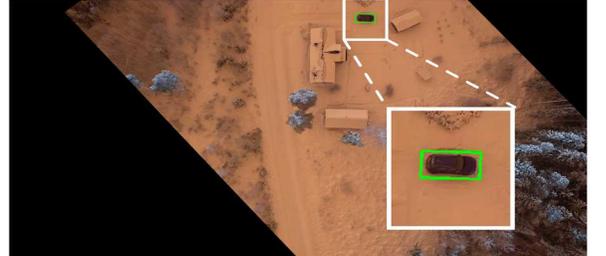

Sample5: altitude (250), Snow cover (Fresh 5-10), cloud cover (clear).
Figure 6. NVD samples.

Table 4. Weather conditions and attributes of videos collected for NVD.

| | Flight Altitude | Snow Cover | Cloud Cover | Resolution | FPS | GSD / Pixel Size | Stabilized |
|---|---|---|---|---|---|---|---|
| 1 | 130-200 m | Minimal (0-1 cm) | Overcast | 1920 x 1080 | 5 | 11.5-17.8 cm | TRUE |
| 2 | 130-200 m | Minimal (0-1 cm) | Overcast | 1920 x 1080 | 25 | 11.5-17.8 cm | FALSE |
| 3 | 130-200 m | Minimal (0-1 cm) | Overcast | 1920 x 1080 | 25 | 11.5-17.8 cm | FALSE |
| 4 | 130 m | Minimal (0-1 cm) | Dense | 1920 x 1080 | 25 | 11.5-17.8 cm | FALSE |
| 5 | 130 m | Minimal (0-1 cm) | Dense | 1920 x 1080 | 25 | 11.5-17.8 cm | FALSE |
| 6 | 250 m | Fresh (1-2 cm) | Light | 1920 x 1080 | 25 | 22.2 cm | FALSE |
| 7 | 250 m | Fresh (1-2 cm) | Light | 1920 x 1080 | 25 | 22.2 cm | FALSE |
| 8 | 250 m | Fresh (5-10 cm) | Clear | 3840 x 2160 | 25 | 11.1 cm | FALSE |
| 9 | 250 m | Fresh (5-10 cm) | Clear | 1920 x 1080 | 5 | 20.2 cm | TRUE |
| 10 | 250 m | Fresh (5-10 cm) | Clear | 3840 x 2160 | 25 | 11.1 cm | FALSE |
| 11 | 250 m | Fresh (5-10 cm) | Clear | 3840 x 2160 | 25 | 11.1 cm | FALSE |
| 12 | 250 m | Fresh (5-10 cm) | Clear | 3840 x 2160 | 25 | 11.1 cm | FALSE |
| 13 | 120 m | No snow | Clear | 1920 x 1080 | 25 | 11.2 cm | FALSE |
| 14 | 250m | Fresh (10-15 cm) | Dense | 1920 x 1080 | 25 | 11.5-17.8 cm | FALSE |
| 15 | 250m | Fresh (10-15 cm) | Dense | 1920 x 1080 | 5 | 11.5-17.8 cm | TRUE |
| 16 | 250 m | Fresh (5-10 cm) | Clear | 1920 x 1080 | 5 | 11.1 cm | TRUE |
| 17 | 130-200 m | Minimal (0-1 cm) | Overcast | 1920 x 1080 | 5 | 11.5-17.8 cm | TRUE |
| 18 | 150 m | Minimal (0-1 cm) | Dense | 1920 x 1080 | 5 | 11.5-17.8 cm | TRUE |
| 19 | 250 m | Fresh (1-2 cm) | Light | 1920 x 1080 | 5 | 22.2 cm | TRUE |
| 20 | 250 m | Fresh (1-2 cm) | Light | 1920 x 1080 | 5 | 22.2 cm | TRUE |
| 21 | 250 m | Fresh (5-10 cm) | Clear | 1920 x 1080 | 5 | 11.1 cm | TRUE |
| 22 | 250 m | Fresh (5-10 cm) | Clear | 1920 x 1080 | 5 | 22.2 cm | TRUE |

## 4. Experimental Results

We incorporated three advanced detectors that are widely used in both academic research and industrial applications.
- a- YOLOv5s
- b- YOLOv8s
- c- Faster R-CNN

We assessed the performance of these detectors using NVD and examined how augmentation methods during the data preparation stage affected their performance. For more information on the augmentation methods used, please refer to section 3.2.

The data has been prepared as follows to train and infer the chosen detectors.

- Total frames = 8450
- Train size = 57%
  - o 4844 frames
  - o 14985 vehicles
- Val. size = 14%
  - o 1212 frames
  - o 3991 vehicles
- Test size = 28%
  - o 2394 frames
  - o 7337 vehicles

The performance of state-of-the-art detectors was measured under different augmentation techniques applied to the NVD dataset in order to assess their impact. The results for the various augmentation techniques used are presented in Tables 5 and 6.

Table 5. Performance of STOA detectors on NVD.

| Model | Precision | Recall | mAP50 | mAP50-95 |
|---|---|---|---|---|
| YOLOv5s | 54.2% | 33.7% | 47.3% | 30.5% |
| YOLOv5s_Au* | 70.6% | 48.2% | 56.0% | 24.1% |
| YOLOv8s | 65.8% | 22.4% | 45.1% | 29.8% |
| YOLOv8s_Au* | 77.1% | 34.6% | 50.7% | 24.1% |

* Au means with augmentation

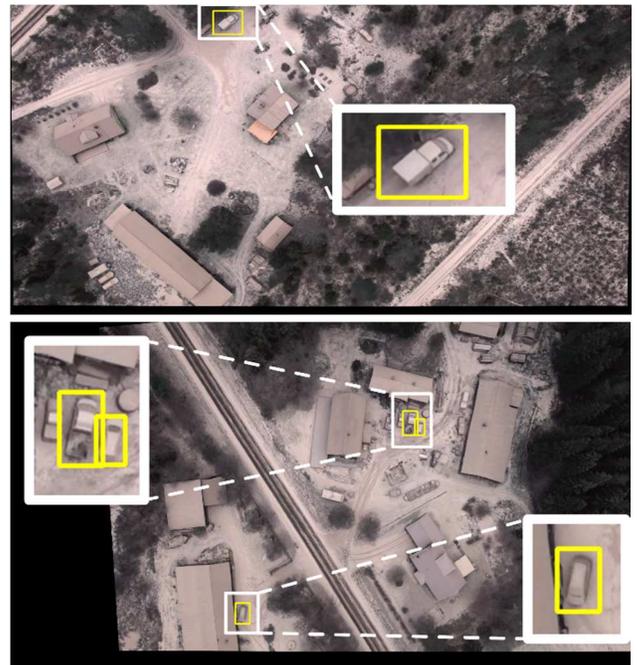

Figure 7. Vehicles detected by YOLOv8s but YOLOv5s.

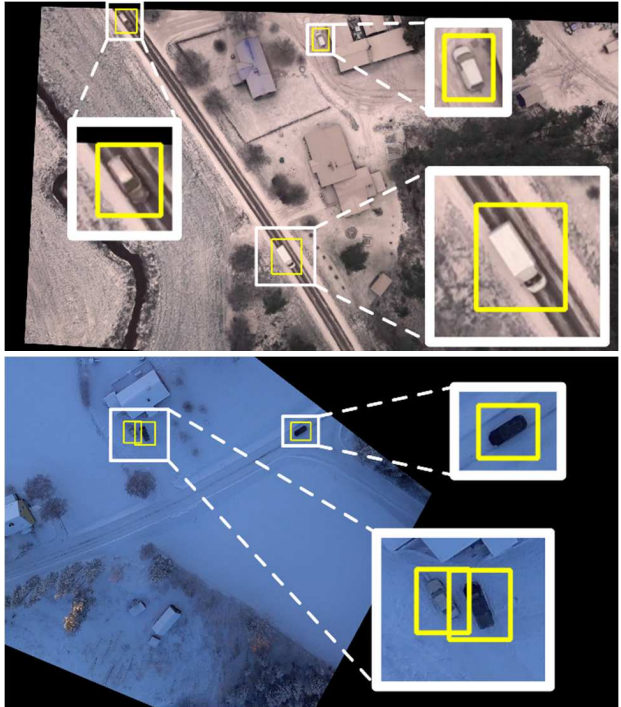

Figure 8. Vehicles detected by YOLOv5s_Au but YOLOv5s.

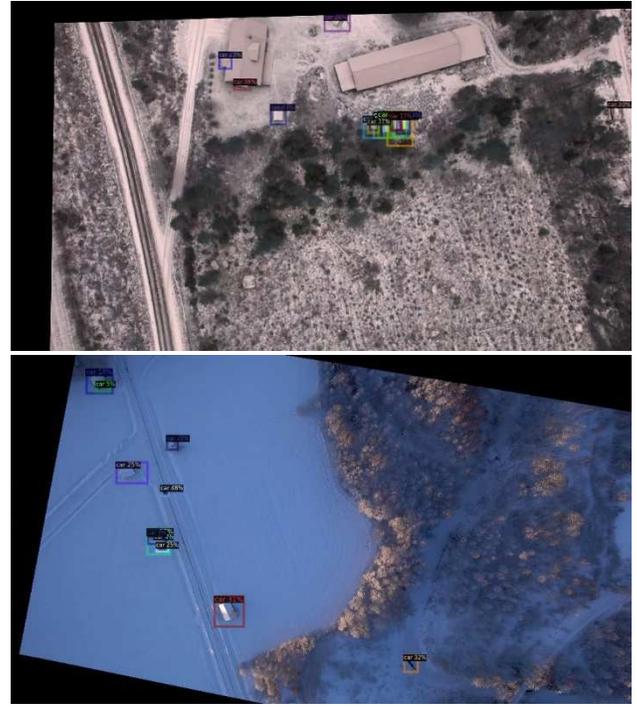

Figure 10. Detection results by Faster RCNN.

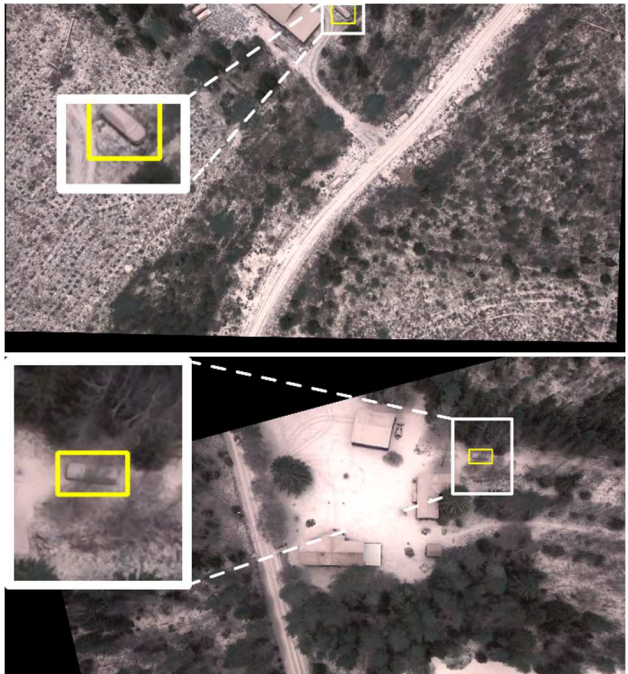

Figure 9. Vehicles detected by YOLOv8s_Au but YOLOv8s.

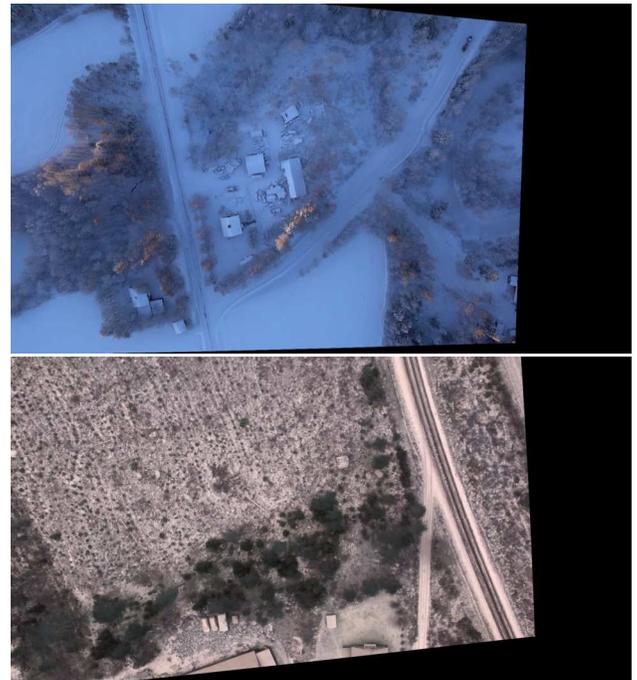

Figure 11. Challenging images - none of the SOTA detectors work.

Table 6. Performance of Faster RCNN on NVD.

| Model | AP | AP50 | AP75 | APs |
|---|---|---|---|---|
| Faster RCNN | 24.428 % | 46.219 % | 23.050% | 35.262 % |

The aim of the following experiment is to assess the performance of the current state-of-the-art detectors in comparison to other available benchmark aerial data. The primary goal is to determine how the performance is affected when using our dataset in contrast to the others.

Table 7. Performance of SOTA detectors on different UAV dataset.

| Dataset | Model | Recall | Precision | AP |
|---|---|---|---|---|
| DLR 3k | Faster RCNN | 78.3 | 89.2 | 79.54 |
| Stanford drone | Faster RCNN | - | - | 75.3 |
| | YOLOv3 | - | - | 80 |
| | YOLOv5s | - | - | 82.5 |
| UAVDT | Faster RCNN | - | - | 27.32 |
| | YOLOv3 | 88.6 | 96.5 | - |
| | YOLOv5s | 90.2 | 97.9 | - |
| CARPK | YOLOv3 | 95 | 97 | - |
| | YOLOv5s | 97.2 | 98.5 | - |
| NVD | Faster RCNN | - | - | 24.4 |
| | YOLOv5s | 54.2% | 33.7% | - |
| | YOLOv8s | 65.8% | 22.4% | - |

## 5. Conclusion

Drones can be used for different purposes related to safety as finding events related to car accidents or traffic congestion, which can be lifesaving. However, the harsh weather conditions and low light during wintertime in the northern hemisphere make search and rescue missions by drones difficult. This work highlights the importance of using adequate training datasets when developing object detectors for drones. The Nordic vehicle dataset (NVD) has been prepared to be used by the research community for better evaluation of the detector performance in varying weather conditions. The results of the experiment show that simply fine-tuning the current state-of-the-art models or augmenting the data will not enable the models to achieve the best possible results. This indicates that there is a need for current research conducted for vehicle detection to utilize and benchmark such challenging data collected in difficult situations. Recently a lot of research has been initiated on removing snow, rain, fog, etc. [30-31], however, the effectiveness of deploying them in real-life snowy conditions with limited computations will be explored in future work.